\documentclass{article}

     \PassOptionsToPackage{numbers}{natbib}

\usepackage[final]{neurips_2023}

\usepackage[utf8]{inputenc} 
\usepackage[T1]{fontenc}    
\usepackage{hyperref}       
\usepackage{url}            
\usepackage{booktabs}       
\usepackage{amsfonts}       
\usepackage{nicefrac}       
\usepackage{microtype}      
\usepackage{xcolor}         
\usepackage{amsmath, amssymb, bm, algorithmic, algorithm2e, xargs, graphicx}

\setcitestyle{square}

\DeclareMathOperator*{\argmin}{argmin} 

\title{
AiRLIHockey: Highly Reactive Contact Control and Stochastic Optimal Shooting
}

\author{%
  Julius Jankowski\thanks{equal contribution} \\
  Idiap Research Institute\\
  Martigny, Switzerland \\
  \texttt{jjankowski@idiap.ch} \\
  \And
  Ante Mari\'c${}^*$ \\
  Idiap Research Institute\\
  Martigny, Switzerland \\
  \texttt{amaric@idiap.ch} \\
  \And
  Sylvain Calinon \\
  Idiap Research Institute\\
  Martigny, Switzerland \\
  \texttt{scalinon@idiap.ch} \\ \\
}

\begin{document}

\maketitle

\begin{abstract}
Air hockey is a highly reactive game which requires the player to quickly reason over stochastic puck and contact dynamics. We implement a hierarchical framework which combines stochastic optimal control for planning shooting angles and sampling-based model-predictive control for continuously generating constrained mallet trajectories. Our agent was deployed and evaluated in simulation and on a physical setup as part of the \textit{Robot Air-Hockey challenge} competition at NeurIPS 2023.
\end{abstract}

\begin{figure*}[th!]
    \centering
    \includegraphics[width=.87\linewidth]{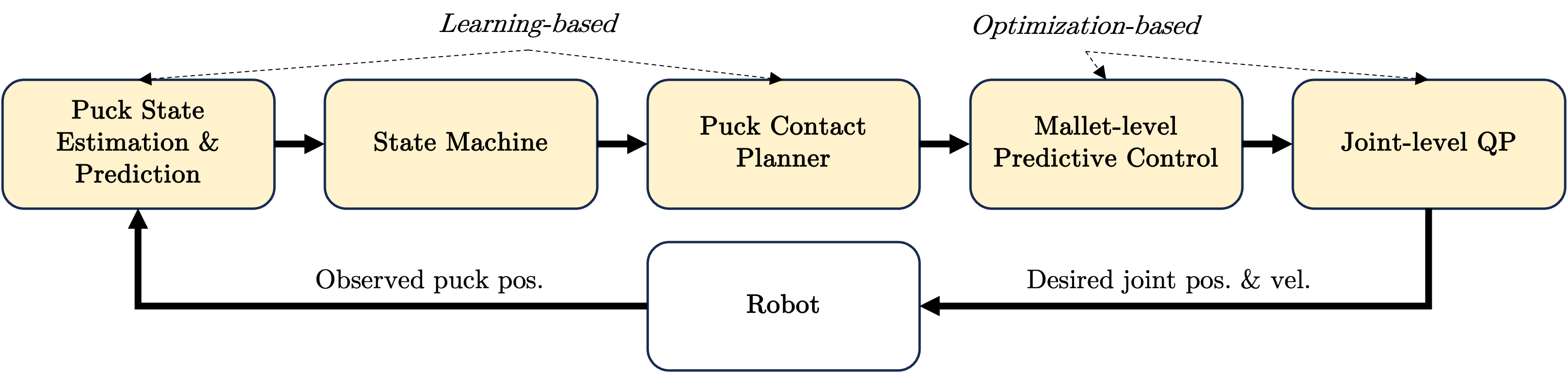}
    \caption{Diagram overview of the \textit{AiRLIHockey} agent. Starting from noisy observations, we estimate the puck state subject to estimated model parameters. Depending on the state, a mode (e.g. shooting or defending) is triggered by a heuristic state machine. The contact state planner and the subsequent mallet controller generate control actions for the mallet. For joint-level control, we use constrained quadratic programming to find the next joint state that tracks the mallet trajectory planned by the higher layers, while staying close to a reference configuration (e.g. a high-manipulability configuration for shooting), with constraints on the joint position, joint velocity, and the z-coordinate of the mallet in order to stay in contact with the table at all times.}
    \label{fig:diagram}
    \vskip -10pt
\end{figure*}

\section{Introduction}
\label{sec:intro}



The fast and stochastic nature of the game \textit{air hockey} requires autonomous systems to reason over contact events in the future at a rate that is sufficient to react to inherent perturbations. The fact that a whole match is to be played without any breaks puts a particular focus on the robustness of the planning and control framework. In this paper, we present our planning and control framework that enables a KUKA iiwa robotic manipulator to play air hockey. While the objective of the game - scoring more goals than the opponent - is obvious, the horizon of a whole match is too long to optimize directly for that objective. Instead, we study primitive skills such as \textit{shooting}, \textit{defending} and \textit{preparing} of the puck. Sequences of these skills can be interpreted as robot policies that aim at maximizing the chance of winning the match. In this work, we focus on the shooting skill and formulate the search for the best shooting angle given a puck state in the future as a stochastic optimal control problem. Since solving such optimization problems is time-consuming and we aim at 50 Hz replanning rates, we propose to solve many of these problems offline and train an energy-based model that represents an implicit shooting angle policy. Another key to the performance of our approach is the subsequent optimization of end-effector (i.e. mallet) trajectories also at a rate of 50 Hz by using zero-order optimization on top of a low-dimensional trajectory representation as in \cite{Jankowski2023}. These trajectories are constrained to keep the mallet on the table without colliding with the walls and to connect the current mallet state with the shooting mallet position that is given by the optimal shooting angle. The objective of this second optimization is to maximize the mallet velocity in shooting direction without violating joint velocity limits.

Fig.~\ref{fig:diagram} illustrates the sub-modules that are part of the framework. Section \ref{sec:state} presents our approach of learning locally linear stochastic puck dynamics. Using this model, we implement a stochastic optimal contact planner detailed in Sections \ref{sec:planner} and \ref{sec:mpc}. The contact planner operates across three different behavior modes - shooting, defending, and preparing for a shot. We use a heuristics-guided state machine to switch between different behavior modes. Section \ref{sec:results} gives an overview of competition results.

\section{Puck State Estimation \& Prediction}
\label{sec:state}

We model the puck dynamics as piecewise (locally) linear with three different modes: \textit{a)} The puck is not in contact with a wall or a mallet. \textit{b)} The puck is in contact with a wall. \textit{c)} The puck is in contact with a mallet. We use data collected in simulation to learn linear parameters $\bm{A}_i, \bm{B}_i$, and an individual covariance matrix $\bm{\Sigma}_i$ representing the process noise for each mode. This gives us a probability distribution over puck trajectories
\begin{equation}
    \mathrm{Pr}_i(\bm{s}_{k+1}^p | \bm{s}_{k}^p, \bm{s}_{k}^m) = \mathcal{N}\Big(\bm{A}_i \bm{s}_{k}^p + \bm{B}_i \bm{s}_{k}^m, \bm{\Sigma}_i(\bm{s}_{k}^p, \bm{s}_{k}^m)\Big),
\end{equation}
with puck state $\bm{s}_{k}^p = (\bm{x}_{k}^p, \dot{\bm{x}}_{k}^p)$ and mallet state $\bm{s}_{k}^m = (\bm{x}_{k}^m, \dot{\bm{x}}_{k}^m)$. While each of the modes is a Gaussian distribution, marginalizing over the mallet state in order to propagate the puck state is not possible due to the discontinuity stemming from contacts. Therefore, we utilize an extended Kalman filter to estimate puck states across different timesteps for any given mallet state.

\section{Stochastic Optimal Shooting Planner}
\label{sec:planner}

Depending on the state machine status, the robot is required to generate a \textit{shooting}, \textit{defending} or \textit{preparing} plan including the robot motion from its current state up until the mallet makes contact with the puck. In all cases, we simplify the planning by separating the problem into two phases: \textit{1)} Generating a stochastic optimal contact state for the mallet and puck, and \textit{2)} Generating an optimal mallet trajectory that connects the current mallet state with the contact mallet state planned in \textit{1)}.


\subsection{Shooting}
\begin{figure}[t]
    \centering
    \includegraphics[width=0.95\linewidth]{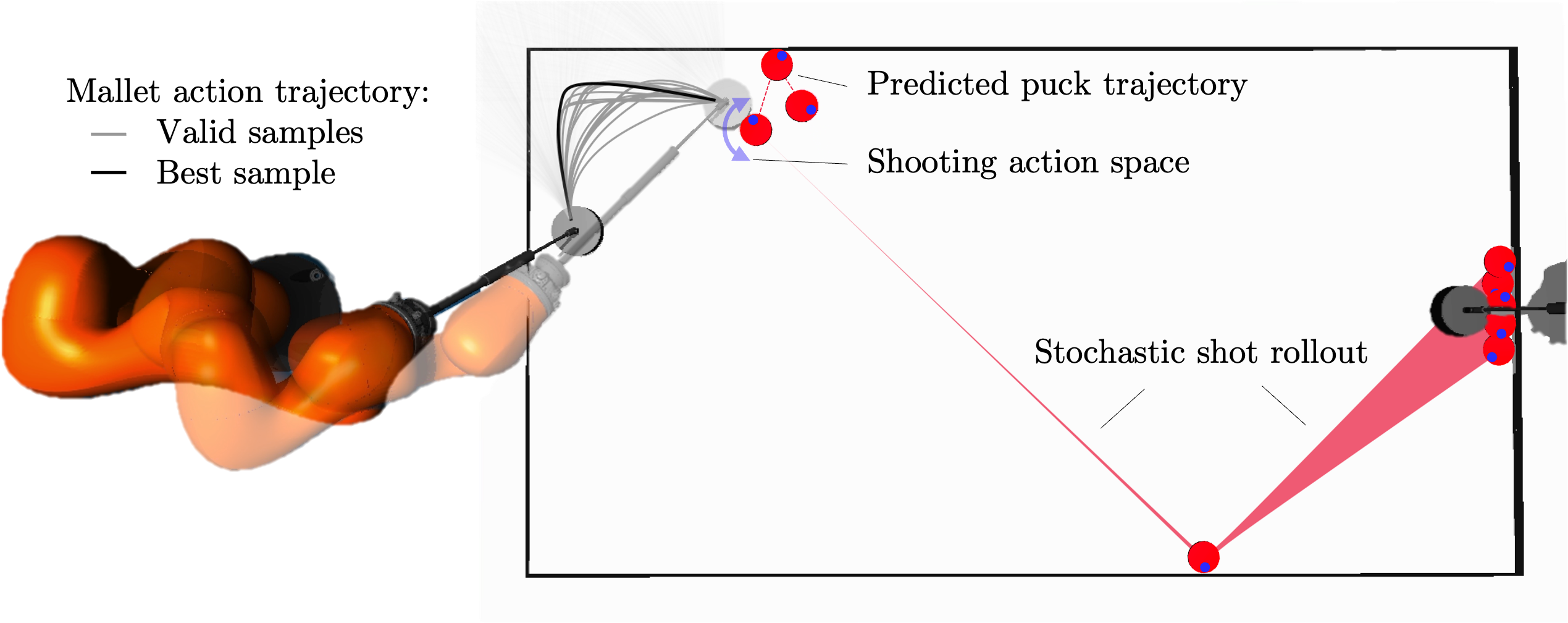}
    \caption{Overview of the interplay between the puck and the mallet for the subtask of scoring a goal. Given the mallet position and the estimated puck position, our framework generates a motion plan for the mallet such that the score probability is maximized.}
    \label{fig:example}
    \vskip -10pt
\end{figure}

\begin{figure}[h!]
    \centering
    \includegraphics[width=0.9\linewidth]{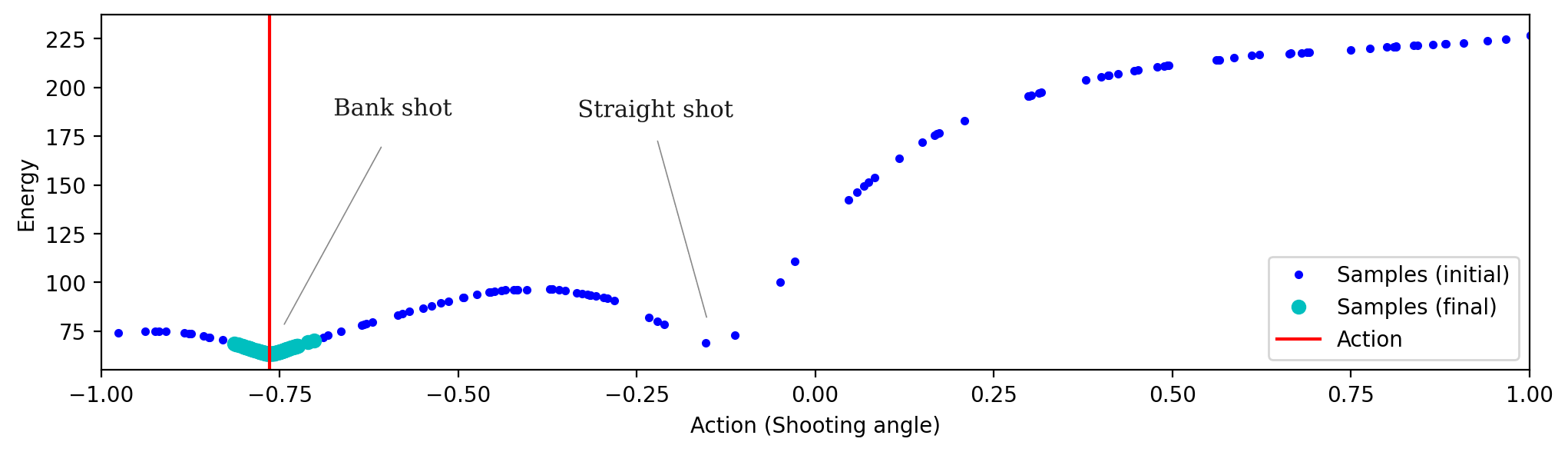}
    \caption{Sampling the action space for an example scenario shown in Figure \ref{fig:example}. Dark blue points show samples at the initial timestep. Iterative recentering and variance reductions lead to the final samples shown in cyan. The red vertical line denotes the selected action.}
    \label{fig:sampling}
    \vskip -10pt
\end{figure}

Due to process and observation noise, we pose the planning of the shooting contact state as a stochastic optimal control problem and simplify it by making the following assumptions:
\begin{enumerate}
    \item The contact between the mallet and the puck always happens exactly at $k=0$.
    We are only interested in a single mallet state, 
    and we can ignore all other mallet states after contact.
    \item The planning horizon $K$, i.e. the time after the contact, is chosen such that the puck state of interest is exactly occurring at $k=K$.
    Thus, we can reduce the computation of the quality metric (the probability of scoring a goal) to the final puck state. 
    \item The time of contact is preset.
    \item The mallet velocity at the time of the contact is always set to be the maximum that the robot can generate for a given mallet position while respecting the dynamic constraints.
\end{enumerate}
Keeping these assumptions in mind, we can formulate a stochastic optimal control problem
\begin{align}
\label{eq:soc}
    &\min_{\bm{s}_{0}^m} J\left(\mathrm{Pr}(\bm{s}_{K}^p)\right), \quad
    ~\mathrm{s.t.} \quad ||\bm{x}_{0}^m - \bm{x}_{0}^p||_2 = r^m + r^p,
\end{align}
in which we aim at minimizing a function of the probability distribution of the puck state. In order to compute the goal scoring cost given a candidate contact state, we approximate the probability distribution over puck trajectories as described in Section \ref{sec:state}. We define the cost 
as a weighted sum of the probability of scoring a goal and the expected puck velocity at the goal line
with an additional penalty term on all trajectories with 
a low probability of entering the goal. 
Different trade-offs between puck velocity and probability of scoring a goal can be achieved by tuning the cost coefficients. 
Lastly, we leverage the puck-mallet contact constraints to reduce our input space to a single dimension.
In order to solve the posed optimization problem at the same control rate of $50$ Hz as the middle-level controller, we transfer the heavy computational burden to an offline phase in which we collect optimal plans for a variety of scenarios of interest. We then use the collected data to train a behavior cloning model to rapidly recover an optimal plan for the online scenario at hand. We use an implicit energy-based model \cite{Florence2022} to clone the behavior of our offline controller by mapping state-action pairs to an energy function. 
We use the trained energy-based model to find the optimal mallet angle $\hat{ a}$ relative to the puck for a given initial puck state $\pmb {s}_{0}^p$:
\begin{equation}
\label{eq:ebm}
    \hat{a}=\argmin_{a\in \mathcal{A}}{E_\theta(\pmb {s}_{0}^p, a)}.
\end{equation}
During runtime we optimize 
by iteratively sampling a number of candidate actions at each timestep with recentering and reductions on the sampling variance. We select an action leading to the lowest energy when paired with the puck state, as shown in Figure \ref{fig:sampling}.

\subsection{Defense \& preparation}

To generate an optimal mallet state at the time of contact for deflecting the puck away from our goal, we apply a sample-based optimization technique online. The objective of the optimization is to achieve a desired vertical puck velocity after contact (e.g. close to zero).
We apply the same sample-based optimization technique to prepare the puck for a shot. The goal here is to move the puck towards the horizontal center by bouncing the puck against the wall first. Since this task does not require high precision, we define again a target puck state after contact as the objective for the optimization problem. The target puck state, i.e. direction and speed, is computed heuristically based on the puck position at the time of the contact.

\section{Mallet-level Model Predictive Control}
\label{sec:mpc}

The mid-level control layer is responsible for making the mallet hit the puck as planned by the contact planner without colliding with the walls of the table.
Thus, the constraint for the mallet controller is to reach a certain mallet state 
at a given point in time. 
We reduce the dimensionality of the decision variable for lower computational burden by using a trajectory parameterization that ensures most of our constraints \cite{Jankowski2023}.
Our parameterization also minimizes an acceleration functional, i.e. the sum over the squared acceleration is minimal. By computing the basis functions of the trajectory parameterization offline, we again transfer a significant part of the computational burden from the online control loops. In order to exert control over the trajectory, we use the final mallet velocity 
as a decision variable and use the error w.r.t. to the desired velocity as the new objective. We find the optimal trajectory by sampling 
a number of candidate mallet velocities 
and evaluating the corresponding rolled-out trajectories (see Fig.~\ref{fig:example}). 
The first mallet action leading to the lowest-cost trajectory is then applied as a control reference.




\section{Competition results}
\label{sec:results}

The tournament stage of the competition evaluated agents on a simulated KUKA iiwa14 LBR manipulator in a double-round robin tournament scheme. Our agent was able to win all matches within the tournament. We ranked first among 7 participants, with scoring based on the accumulated number of wins, draws, and losses. 
For detailed results and scoring information, we refer the reader to the competition website: \url{https://air-hockey-challenge.robot-learning.net/}.



\section{Conclusion}

This paper investigates the viability of combining a learned planner with sampling-based model-predictive control in order to achieve reactive behavior. We tested and validated our approach in simulation during the \textit{Air Hockey challenge}, where it achieved best performance among the competitors. Future work will investigate the challenges of transferring our agent to a physical system.

\begin{ack}
We thank the organizers for initiating and providing support during the challenge. This work was supported in part by the Swiss National Science Foundation (SNSF) through the CODIMAN project, and by the State Secretariat for Education,
Research and Innovation in Switzerland for participation in the European Commission’s Horizon Europe Program through the INTELLIMAN project (https://intelliman-project.eu/, HORIZON-CL4-Digital-Emerging Grant 101070136) and the SESTOSENSO project (http://sestosenso.eu/, HORIZON-CL4-Digital-Emerging Grant 101070310).
\end{ack}


\bibliographystyle{ieeetr}
\bibliography{main}

\begin{thebibliography}{1}

\bibitem{Jankowski2023}
J.~Jankowski, L.~Brudermüller, N.~Hawes, and S.~Calinon, ``Vp-sto: Via-point-based stochastic trajectory optimization for reactive robot behavior,'' in {\em 2023 IEEE International Conference on Robotics and Automation (ICRA)}, pp.~10125--10131, 2023.

\bibitem{Florence2022}
P.~Florence, C.~Lynch, A.~Zeng, O.~A. Ramirez, A.~Wahid, L.~Downs, A.~Wong, J.~Lee, I.~Mordatch, and J.~Tompson, ``Implicit behavioral cloning,'' in {\em Proceedings of the 5th Conference on Robot Learning}, vol.~164 of {\em Proceedings of Machine Learning Research}, pp.~158--168, 2022.

\end{thebibliography}


\end{document}